\documentstyle[]{article}

\newcounter{examplectr}
\newcounter{subexamplectr}

\newenvironment{ex}%
   {\vspace{.1in}\addtocounter{examplectr}{1}
     \setcounter{subexamplectr}{0}
     \begin{list}
       {(\arabic{examplectr})}%
       {\setlength{\topsep}{0in}
        \setlength{\leftmargin}{.25in}
               \setlength{\labelsep}{0.075in}}
       \item \begin{minipage}[t]{13.5cm} %was 3.5in
   }%
   {\end{minipage}
    \end{list}\vspace{.1in}}
\newenvironment{subex}%
   { \addtocounter{subexamplectr}{1}
     \begin{list}
       {\alph{subexamplectr}}%
       {\setlength{\topsep}{-\parskip}
        \setlength{\leftmargin}{0.175in}
        \setlength{\labelsep}{0.075in}}
       \item
   }%
   {\end{list}}
\newcommand{\exnum}[2]{\addtocounter{examplectr}{#1}(\arabic{examplectr}{#2})\addtocounter{examplectr}{-#1}}
\newcommand {\etal} {{\it et al.}}
\newcommand{\ssp}{\setlength{\baselineskip}{14.5pt}}
\begin{document}
\pagestyle{empty}
\ssp
\setcounter{secnumdepth}{2}
\noindent
{\it In Proceedings of the ACL/SIGPARSE 4th International
Workshop on Parsing Technologies, Prague / Karlovy Vary, Czech Republic,
1995, pp 48--58.}
\vspace{3mm}
\begin{center}
\begingroup
\def\thefootnote{\fnsymbol{footnote}}
{\large\bf DEVELOPING AND EVALUATING A PROBABILISTIC LR\\[1mm]
PARSER OF PART-OF-SPEECH AND PUNCTUATION LABELS\footnotemark\\[6mm]
\footnotetext{Some of this work was
carried out while the first author was visting Rank Xerox, Grenoble. The work
was also supported by DTI/SALT project 41/5808 `Integrated Language
Database'. Geoff Nunberg provided encouragement and much advice on the
analysis of punctuation, and Greg Grefenstette undertook the original
tokenisation and segmentation of Susanne. Bernie Jones and Kiku Ribas made
helpful comments on an earlier draft. We are responsible for any
mistakes.}}
\endgroup
\setcounter{footnote}{0}
{\large Ted Briscoe\\[3mm]
John Carroll}\\[1mm]
Computer Laboratory, University of Cambridge \\
Pembroke Street, Cambridge CB2 3QG, UK \\[1mm]
ejb / jac@cl.cam.ac.uk
\vspace{3mm}
\def\abstractname{ }
\begin{abstract}
\noindent
We describe an approach to robust domain-independent syntactic parsing
of unrestricted naturally-occurring (English) input. The technique involves
parsing sequences of part-of-speech and punctuation labels using a
unification-based grammar coupled with a probabilistic LR parser. We describe
the coverage of several corpora using this grammar and report the results of a
parsing experiment using probabilities derived from bracketed training data.
We report the first substantial experiments to assess the contribution of
punctuation to deriving an accurate syntactic analysis, by parsing identical
texts both with and without naturally-occurring punctuation marks.
\end{abstract}
\end{center}

\section{Introduction}

This work is part of an effort to develop a robust, domain-independent
syntactic parser capable of yielding the one correct analysis for
unrestricted naturally-occurring input. Our goal is to develop a
system with performance comparable to extant part-of-speech taggers,
returning a syntactic analysis from which predicate-argument structure
can be recovered, and which can support semantic interpretation. The
requirement for a domain-independent analyser favours statistical
techniques to resolve ambiguities, whilst the latter goal favours a
more sophisticated grammatical formalism than is typical in
statistical approaches to robust analysis of corpus material.

Briscoe and Carroll (1993) describe a probablistic parser using a
wide-coverage uni\-fication-based grammar of English written in the
Alvey Natural Language Tools (ANLT) metagrammatical formalism (Briscoe
\etal, 1987), generating around 800 rules in a syntactic
variant of the Definite Clause Grammar formalism (DCG, Pereira \&
Warren, 1980) extended with iterative (Kleene) operators. The ANLT grammar is
linked to a lexicon containing about 64K entries for 40K lexemes,
including detailed subcategorisation information appropriate for the
grammar, built semi-automatically from a learners' dictionary (Carroll
\& Grover, 1989).  The resulting parser is efficient, capable of
constructing a parse forest in what seems to be roughly quadratic time, and
efficiently returning the ranked {\it n}-most likely analyses (Carroll, 1993,
1994).  The probabilistic model is a refinement of probabilistic context-free
grammar (PCFG) conditioning CF `backbone' rule application on LR state
and lookahead item. Unification of the `residue' of features not
incorporated into the backbone is performed at parse time in
conjunction with reduce operations. Unification failure results in the
associated derivation being assigned
a probability of zero. Probabilities are assigned to transitions in
the LALR(1) action table via a process of supervised training based on
computing the frequency with which transitions are traversed in a
corpus of parse histories. The result is a probabilistic parser which,
unlike a PCFG, is capable of probabilistically discriminating
derivations which differ only in terms of order of application of the
same set of CF backbone rules, due to the parse context defined by the
LR table.

Experiments with this system revealed three major problems which our
current research is addressing.  Firstly, although the system is able to
rank parses with a 75\% chance that the correct analysis will be the
most highly ranked, further improvement will require a `lexicalised'
system in which (minimally) probabilities are associated with
alternative subcategorisation possibilities of individual lexical
items. Currently, the relative frequency of subcategorisation
possibilities for individual lexical items is not recorded in
wide-coverage lexicons, such as ANLT or COMLEX (Grishman \etal, 1994).
Secondly, removal of punctuation from the input (after segmentation
into text sentences) worsens performance as punctuation both reduces
syntactic ambiguity (Jones, 1994) and signals non-syntactic
(discourse) relations between text units (Nunberg, 1990). Thirdly, the
largest source of error on unseen input is the omission of appropriate
subcategorisation values for lexical items (mostly verbs), preventing the
system from finding the correct analysis. The current coverage of this system
on a general corpus (e.g.\ Brown or LOB) is estimated to be around 20\% by
Briscoe (1994). We have developed a variant probabilistic LR parser which does
not rely on subcategorisation and uses punctuation to reduce ambiguity.
The analyses produced by this parser could be utilised for phrase-finding
applications, recovery of subcategorisation frames, and other `intermediate'
level parsing problems.

\section{Part-of-speech Tag Sequence Grammar}

Several robust parsing systems exploit the comparative success of
part-of-speech (PoS) taggers, such as Fidditch (Hindle, 1989) or MITFP
(de Marcken, 1990), by reducing the input to a determinate sequence of
extended PoS labels of the type which can be practically disambiguated
in context using a (H)MM PoS tagger (e.g.\ Church, 1988). Such approaches,
by definition, cannot exploit subcategorisation, and probably achieve
some of their robustness as a result. However, such parsers typically
also employ heuristic rules, such as `low' attachment of PPs to
produce unique `canonical' analyses. This latter step complicates the
recovery of predicate-argument structure and does not integrate with a
probabilistic approach to parsing.

We utilised the ANLT metagrammatical formalism to develop a
feature-based, declarative description of PoS label sequences for
English. This grammar compiles into a DCG-like grammar of
approximately 400 rules. It has been designed to enumerate possible
valencies for predicates (verbs, adjectives and nouns) by including
separate rules for each pattern of possible complementation in
English. The distinction between arguments and adjuncts is expressed,
following X-bar theory (e.g.\ Jackendoff, 1977), by Chomsky-adjunction
of adjuncts to maximal projections (\mbox{XP $\rightarrow$ XP
Adjunct}) as opposed to government of arguments (i.e. arguments are
sisters within X1 projections; \mbox{X1 $\rightarrow$ X0 Arg1$\ldots$
ArgN}).  Although the grammar enumerates complementation
possibilities and checks for global sentential well-formedness, it is
best described as `intermediate' as it does not attempt to associate
`displaced' constituents with their canonical position / grammatical
role.

The other difference between this grammar and a more conventional one
is that it incorporates some rules specifically designed to overcome
limitations or idiosyncrasies of the PoS tagging process. For example,
past particles functioning adjectivally, as in {\it The disembodied head}, are
frequently tagged as past participles (VVN) i.e.\ {\it The\_AT disembodied\_VVN
head\_NN1}, so the grammar incorporates a rule which parses
past participles as adjectival premodifiers in this context.
Similar idiosyncratic rules are incorporated for dealing with gerunds,
adjective-noun conversions, idiom sequences, and so forth.

This grammar was developed and refined in a
corpus-based fashion (e.g.\ see Black, 1993) by testing against sentences
from the Susanne corpus (Sampson, 1994), a 138K word treebanked and balanced
subset of the Brown corpus\footnote{The grammar currently covers more than 75\%
of the sentences. Many of the remaining failures for shorter text
sentences are a consequence of the root S node requirement, since they
represent elliptical noun or prepositional phrases in dialogue.  Other failures
on sentences are a consequence of incorporation of complementation
constraints for auxiliary verbs into the grammar but the lack of any
treatment of unbounded dependencies. Nevertheless, we tolerate these
`deficiencies', since they have the effect of limiting the number of
analyses recovered in other cases, and will not, for example, affect
unduly the recovery of subcategorisation frames from the resulting analyses.}.

\section{Text Grammar and Punctuation}

Nunberg (1990) develops a partial `text' grammar for English
which incorporates many constraints that (ultimately) restrict
syntactic and semantic interpretation. For example, textual adjunct
clauses introduced by colons scope over following punctuation, as
\exnum{+1}{a} illustrates; whilst textual adjuncts introduced by
dashes cannot intervene between a bracketed adjunct and the textual
unit to which it attaches, as in \exnum{+1}{b}.
\begin{ex}
\begin{subex}
*He told them his reason: he would not renegotiate his contract,  but
he did not explain to the team owners. (vs. but would stay)
\end{subex}
\begin{subex}
*She left --  who could blame her -- (during the chainsaw scene) and went home.
\end{subex}
\end{ex}

We have developed a declarative grammar in the ANLT metagrammatical
formalism, based on Nunberg's procedural description.  This grammar
captures the bulk of the text-sentential constraints described by
Nunberg with a grammar which compiles into 26 DCG-like
rules. Text
grammar analyses are useful because they demarcate some of the
syntactic boundaries in the text sentence and thus reduce ambiguity,
and because they identify the units for which a syntactic analysis
should, in principle, be found; for example, in \exnum{+1}{}, the
absence of dashes would mislead a parser into seeking a syntactic
relationship between {\it three} and the following names, whilst in
fact there is only a discourse relation of elaboration between this
text adjunct and pronominal {\it three}.
\begin{ex}
The three -- Miles J. Cooperman, Sheldon
Teller, and Richard Austin -- and eight other defendants were charged
in six indictments with conspiracy to violate federal narcotic law.
\end{ex}

The rules of the text grammar divide into three groups: those
introducing text-sentences, those defining text adjunct introduction
and those defining text adjuncts (Nunberg, 1990). An example of each
type of rule is given in \exnum{+1}{a--c}.
\begin{ex}
\begin{subex}
T/txt-sc1 : TxtS $\rightarrow$ (Tu[+sc])* Tu[-sc]
(+pex$\mid$+pqu)
\end{subex}
\begin{subex}
Ta/dash- : Tu[-sc] $\rightarrow$ T[-sc, -cl, -da] Ta[+da,
da-]
\end{subex}
\begin{subex}
T/t\_ta-da-\_t : Ta[+da, da-] $\rightarrow$ +pda Tu[-sc,
-da]
\end{subex}
\end{ex}
These rules are phrase structure schemata employing iterative
operators, optionality and disjunction, preceded by a mnemonic name.
Non-terminal categories are text sentences, units or adjuncts which
carry features mostly representing the punctuation marks which occur
as daughters in the rules (e.g.\ +sc represents presence of a
semi-colon marker), whilst terminal punctuation is represented as +pxx
(e.g.\ +pda, dash). \exnum{+0}{a} states that a text sentence can contain zero
or more text units with a semi-colon at their right boundary followed by
a text unit optionally followed by a question or exclamation
mark. \exnum{+0}{b} states that a text unit not containing a semi-colon can
consist of a text unit or adjunct not containing dashes, colons or
semi-colons followed by a text adjunct introduced by a dash. This type
of `unbalanced' adjunct can only be expanded by \exnum{+0}{c} which states
that it consists of a single opening dash followed by a text unit
which does not itself contain dashes or semi-colons. The features on
the first daughter of \exnum{+0}{b} force dash adjuncts to have lower
precedence and narrower scope than colons or semi-colons, blocking
interpretations of multiple dashes as sequences of `unbalanced'
adjuncts.

Nunberg (1990) invokes rules of (point) absorption which delete
punctuation marks (inserted according to a simple context-free text
grammar) when adjacent to other `stronger' punctuation marks. For
instance, he treats all dash interpolated text adjuncts as
underlyingly balanced, but allows a rule of point absorption to
convert \exnum{+1}{a} into \exnum{+1}{b}.
\begin{ex}
\begin{subex}
*Max fell -- John had kicked him --.
\end{subex}
\begin{subex}
Max fell -- John had kicked him.
\end{subex}
\end{ex}
The various rules of absorption introduce procedurality into the
grammatical framework and require the positing of underlying forms
which are not attested in text. For this reason, `absorption' effects
are captured through propagation of featural constraints in parse
trees. For instance, \exnum{0}{a} is blocked by including distinct
rules for the introduction of balanced and unbalanced text adjuncts
and only licensing the latter text sentence finally.

The text grammar has been tested on Susanne and covers 99.8\% of
sentences. (The failures are mostly text segmentation problems). The
number of analyses varies from one (71\%) to the thousands (0.1\%).
Just over 50\% of Susanne sentences contain some punctuation, so
around 20\% of the singleton parses are punctuated. The major source
of ambiguity in the analysis of punctuation concerns the function of
commas and their relative scope as a result of a decision to
distinguish delimiters and separators (Nunberg 1990:36). Therefore, a
text sentence containing eight commas (and no other punctuation) will
have 3170 analyses. The multiple uses of commas cannot be resolved
without access to (at least) the syntactic context of occurrence.

\section{The Integrated Grammar}

Despite Nunberg's observation that text grammar is distinct from
syntax, text grammatical ambiguity favours interleaved application of
text grammatical and syntactic constraints.  The integration of text
and PoS sequence grammars is straightforward and remains modular, in
that the text grammar is `folded into' the PoS sequence grammar, by
treating text and syntactic categories as overlapping and dealing with
the properties of each using disjoint sets of features, principles of
feature propagation, and so forth. The text grammar rules are
represented as left or right branching rules of `Chomsky-adjunction'
to lexical or phrasal constituents. For example, the simplified rule
for combining NP appositional or parenthetical text adjuncts is
\mbox{N2[+ta] $\rightarrow$ H2 Ta[+bal]} which states that a NP
containing a textual adjunct consists of a head NP followed by a
textual adjunct with balanced delimiters (dashes, brackets or
commas). Rules of this form ensure that syntactic and textual analysis
are mutually `transparent' and orthogonal so, for example, any rules
of semantic interpretation associated with syntactic rules continue to
function unmodified. Such rules attach text adjuncts to the
constituents over which they semantically scope, so it would be
possible, in principle, to develop a semantics for them.  In addition
to the core text grammatical rules which carry over unchanged from the
stand-alone text grammar, 44 syntactic rules (of pre- and post-
posing, and coordination) now include (often optional) comma markers
corresponding to the purely `syntactic' uses of punctuation.

The approach to text grammar taken here is in many ways similar to
that of Jones (1994). However, he opts to treat punctuation marks as
clitics on words which introduce additional featural information into
standard syntactic rules. Thus, his grammar is thoroughly integrated
and it would be harder to extract an independent text grammar or build
a modular semantics. The coverage of the integrated version of the
text grammar is described in more detail in Briscoe \& Carroll
(1994).

\section{Parsing the Susanne and SEC Corpora}

The integrated grammar has been used to parse Susanne and the quite
distinct SEC Corpus (Taylor \& Knowles, 1988), a 50K word treebanked
corpus of transcribed British radio programmes punctuated by the
corpus compilers. Both corpora were retagged with determinate
punctuation and PoS labelling using the Acquilex HMM tagger (Elworthy,
1993, 1994) trained on text tagged with a slightly modified version of
CLAWS-II labels (Garside \etal, 1987).

\subsection{Coverage and Average Ambiguity}

To examine the efficiency and coverage of the grammar we applied it to
our retagged versions of Susanne and SEC. We used the ANLT chart
parser (Carroll, 1993), but modified just to count the number of
possible parses in the parse forests (Billot \& Lang, 1989) rather
than actually unpacking them. We also imposed a per-sentence time-out
of 30 seconds CPU time, running in Franz Allegro Common Lisp 4.2 on an
HP PA-RISC 715/100 workstation with 96 Mbytes of physical memory.

We define the `coverage' of the grammar to be the inverse of the
proportion of sentences for which no analysis was found---a weak measure
since discovery of one or more global analyses does not entail that the
correct analysis is recovered. For both corpora, the majority of sentences
analysed successfully received under 100 parses, although there is a long tail
in the distribution.  Monitoring this distribution is helpful during
grammar development to ensure that coverage is increasing but the
ambiguity rate is not. A more succinct though less intuitive measure
of ambiguity rate for a given corpus is what we call the average
parse base (APB), defined as the geometric mean over all sentences in the
corpus of $\sqrt[n]{p}$, where $n$ is the number of words in a sentence, and
$p$, the number of parses for that sentence\footnote{Black
\etal (1993:13) define an apparently similar measure, {\it parse base}, as the
``geometric mean of the number of parses per word for the entire corpus'', but
in the immediately following sentence talk about raising it to the power of
the number of words in a sentence, which is inappropriate for a simple
ratio.}. Thus, given a sentence $n$ tokens long, the
APB raised to the $n$th power gives the number of analyses
that the grammar can be expected to assigned to a sentence of
that length in the corpus. Table
\ref{sus-sec} gives these measures for all of the sentences in Susanne and in
SEC.

\begin{table}
\begin{center}
\begin{tabular}{|l|rr|rr|} \hline
                   & Susanne       && SEC & \\ \hline

Parse fails        & 1745 & 24.9\%  &  898 & 33.1\% \\

1--9 parses        & 1566 & 22.3\%  &  607 & 22.3\% \\

10--99 parses      & 1306 & 18.6\%  &  418 & 15.4\% \\

100--999 parses    &  893 & 12.7\%  &  299 & 11.0\% \\

1K--9.9K parses    &  611 &  8.7\%  &  197 &  7.3\% \\

10K--99K parses    &  413 &  5.9\%  &  108 &  4.0\% \\

100K+ parses       &  475 &  6.8\%  &  189 &  7.0\% \\

Time-outs          &   5 &  0.07\%  &    1 &  0.04\% \\
\hline
Number of sentences        & 7014  &&  2717 & \\

Mean sentence length (MSL)    & 20.1  &&  22.6 & \\

MSL -- fails               & 21.7  &&  27.6 & \\

MSL -- time-outs           & 67.2  &&  79.0 & \\

Average Parse Base         & 1.256  &&  1.239 & \\ \hline
\end{tabular}

\caption{Grammar coverage on Susanne and SEC}
\label{sus-sec}
\end{center}
\end{table}

As the grammar was developed solely with reference to Susanne,
coverage of SEC is quite robust. The two corpora differ considerably
since the former is drawn from American written text whilst the latter
represents British transcribed spoken material. The corpora overall contain
material drawn from widely disparate genres / registers, and are more
complex than those used in DARPA ATIS tests and more diverse than those used
in MUC. The APBs for Susanne and SEC of 1.256
and 1.239 respectively indicate that sentences of average
length in each corpus could be expected to be assigned of the order of 97 and
126 analyses (i.e.\ $1.256^{20.1}$ and $1.239^{22.6}$). Black \etal (1993:156)
quote a {\it parse base} of 1.35 for the IBM grammar for computer manuals
applied to sentences 1--17 words long. Although, as mentioned above, Black's
measure may not be exactly the same as our APB measure, it is probable that
the IBM grammar assigns more analyses than ours for sentences of the same
length. Black achieves a coverage of around 95\%, as opposed to our coverage
rate of 67--74\% on much more heterogeneous data and longer sentences.

The parser throughput on these tests, for sentences successfully
analysed, is around 45 words per CPU second on an HP PA-RISC
715/100. Sentences of up to 30 tokens (words plus punctuation) are
parsed in an average under 0.6 seconds each, whilst those around 60
tokens take on average 4.5 seconds. Nevertheless, the relationship
between sentence length and processing time is fitted well by a
quadratic function, supporting the findings of Carroll (1994) that in
practice NL grammars do not evince worst-case parsing complexity.

\subsubsection{Coverage, Ambiguity and Punctuation}

We have also run experiments to
evaluate the degree to which punctuation is contributing useful
information. Intuitively, we would expect the exploitation of text
grammatical constraints to both reduce ambiguity and extend coverage
(where punctuation cues discourse rather than syntactic relations between
constituents). Jones (1994) reports a preliminary experiment evaluating
reduction of ambiguity by punctuation. However, the grammar he uses was
developed only to cover the test sentences, drawn entirely from
the SEC corpus which was punctuated post hoc by the corpus developers
(Taylor and Knowles, 1988).

We took all in-coverage sentences from Susanne
of length 8--40 words inclusive containing internal punctuation; a total
of 2449 sentences. The APB for this set was 1.273, mean length
22.5 words, giving an expected number of analyses for an average
sentence of 225. We then removed all sentence-internal punctuation from this
set and re-parsed it. Around 8\% of sentences now failed to receive an
analysis. For those that did (mean length 20.7 words), the APB was now
1.320, so an average sentence would be assigned 310 analyses, 38\% more than
before. On closer inspection, the increase in ambiguity is due to two
factors: a) a significant proportion of sentences that previously received
1--9 analyses now receive more, and b) there is a much more substantial tail
in the distribution of sentence length vs.\ number of parses, due to some
longer sentences being assigned many more parses. Manual examination of 100
depunctuated examples revealed that in around a third of cases, although the
system returned global analyses, the correct one was not in this set (Briscoe
\& Carroll, 1994).  With a more constrained (subcategorised) syntactic
grammar, many of these examples would not have received any global
syntactic analysis.

\subsection{Parse Selection}

A probabilistic LR parser was trained with the integrated grammar by
exploiting the Susanne treebank bracketing. An LR parser (Briscoe and
Carroll, 1993) was applied to unlabelled bracketed sentences from the
Susanne treebank, and a new treebank of 1758 correct and complete analyses
with respect to the integrated grammar was constructed semi-automatically by
manually resolving the remaining ambiguities. 250 sentences from the new
treebank were kept back for testing. The remainder, together with a further
set of analyses from 2285 treebank sentences that were not checked manually,
were used to train a probabilistic version of the LR parser, using
Good-Turing smoothing to estimate the probability of unseen transitions in
the LALR(1) table (Briscoe and Carroll, 1993; Carroll, 1993).  The
probabilistic parser can then return a ranking of all possible analyses for a
sentence, or efficiently return just the {\it n}-most probable (Carroll,
1993).

The probabilistic parser was tested on the 250 sentences held out from the
manually-created treebank (with mean length 18.2 tokens, mean number of parses
per sentence 977, and APB 1.252); in this test 85 sentences (34\%) had the
correct analysis ranked in the top three\footnote{This is a strong
measure, since it not only accounts for structural identity between
trees, but also correct rule application at every node.}. This
figure rose to 51\% for sentences of less than 20 words. Considering
just the highest ranked analysis for each sentence, in Sampson, Haigh
\& Atwell's (1989) measure of correct rule application the parser
scored a mean of 83.5\% correct over all 250 sentences. Table \ref{sus-eval}
shows the results of this test---with respect to the original Susanne
bracketings---using the Grammar Evaluation Interest Group scheme (GEIG, see
e.g.\ Harrison \etal, 1991). This compares unlabelled bracketings
derived from corpus treebanks with those derived from parses for the same
sentences by computing {\it recall}, the ratio of matched brackets over all
brackets in the treebank; {\it precision}, the ratio of matched brackets over
all brackets found by the parser; `crossing' brackets, the number of times
a bracketed sequence output by the parser overlaps with one from the
treebank but neither is properly contained in the other; and {\it
minC}, the number of sentences for which all of the analyses had one
or more crossings.
\begin{table}
\begin{center}
\begin{tabular}{|l|rrrr|} \hline

       & minC      & Crossings     & Recall (\%)   & Precision (\%) \\
\hline

{\em Probabilistic parser analyses} &&&& \\
Top-ranked 3 analyses, weighted =
       & 150       & 2.62          & 76.47         & 42.35 \\
Random 3 analyses, weighted =
       & 155       & 3.87          & 67.05         & 37.40 \\
&&&& \\ \hline

{\em Manually-disambiguated analyses} &&&& \\
Single analysis
       & 91        & 0.88          & 91.51         & 50.73 \\ \hline
\end{tabular}

\caption{GEIG evaluation metrics for test set of 250 unseen sentences
(lengths 3--56 words, mean length 18.2)}
\label{sus-eval}
\end{center}
\end{table}
The table also gives an indication of the best and worst possible
performance of the disambiguation component of the system, showing the
results obtained when parse selection is replaced by a simple random
choice, and the results of evaluating the manually-created treebank
against the corresponding Susanne bracketings. In this latter figure, the mean
number of crossings is greater than zero mainly because of compound noun
bracketing ambiguity which our grammar does not attempt to resolve,
always returning a right-branching binary analysis.

Black (1993:7) uses the crossing brackets measure to define a notion
of structural consistency, where the structural consistency rate for
the grammar is defined as the proportion of sentences for which at
least one analysis contains no crossing brackets, and reports a rate
of around 95\% for the IBM grammar tested on the computer manual
corpus. The problem with the GEIG scheme and with structural
consistency is that both are still weak measures (designed to avoid
problems of parser/treebank representational compatibility) which lead
to unintuitive numbers whose significance still depends heavily on
details of the relationship between the representations compared
(c.f.\ the compound noun issue mentioned above).

Schabes \etal\ (1993) and Magerman (1995) report results using the GEIG
evaluation scheme which are numerically superior to ours. However, their
experiments are not strictly compatible because they both utilise
more homogeneous and probably simpler corpora. In addition, Schabes \etal\
do not recover tree labelling, whilst Magerman has developed a parser
designed to produce identical analyses to those used in the Penn Treebank,
removing the problem of spurious errors due to grammatical incompatibility.
Both these approaches achieve better coverage by constructing the grammar
fully automatically. No one has yet shown that any robust parser
is practical and useful for some NLP task. However, it seems likely that
say rule-to-rule semantic interpretation will be easier with hand-constructed
grammars with an explicit, determinate ruleset.
A more meaningful comparison will require application
of different parsers to an identical and extended test suite and
utilisation of a more stringent standard evaluation procedure sensitive to
node labellings.

\subsubsection{Parse Selection and Punctuation}

In order to assess the contribution of punctuation to the selection of
the correct analysis, we applied the same trained version of the
integrated grammar to the 106 sentences from the test set which
contain internal punctuation, both with and without the punctuation
marks in the input. A comparison of the GEIG evaluation metrics for
this set of sentences punctuated and unpunctuated gives a measure of
the contribution of punctuation to parse selection on this data. (The
results for the unpunctuated set were computed against a version of
the Susanne treebank from which punctuation had also been removed.) As
table~\ref{sus-punct} shows, recall declines by 10\%, precision by 5\%
and there are an average of 1.27 more crossing brackets per sentence.
These results indicate clearly that punctuation and text grammatical
constraints can play an important role in parse selection.

\begin{table}
\begin{center}
\begin{tabular}{|l|rrrr|} \hline

       & minC      & Crossings     & Recall (\%)   & Precision (\%) \\
\hline

{\em With punctuation} &&&& \\
Top-ranked 3 analyses, weighted =
       & 78        & 3.25          & 74.38         & 40.78 \\
&&&& \\ \hline

{\em Punctuation removed} &&&& \\
Top-ranked 3 analyses, weighted =
       & 82        & 4.52          & 65.54         & 35.95 \\ \hline
\end{tabular}

\caption{GEIG evaluation metrics for test set of 106 unseen
punctuated sentences (mean length with punctuation 21.4 words; without,
19.6)}
\label{sus-punct}
\end{center}
\end{table}

\section{Conclusions}

Briscoe and Carroll (1993) and Carroll (1993) showed that the LR
model, combined with a grammar exploiting subcategorisation constraints,
could achieve good parse selection accuracy but at the expense of
poor coverage of free text. The results reported here suggest that improved
coverage of heterogeneous text can be achieved by exploiting textual and
grammatical constraints on PoS and punctuation sequences. The
experiments show that grammatical coverage can be
greatly increased by relaxing subcategorisation constraints, and that
text grammatical or punctuation-cued constraints can reduce
ambiguity and increase coverage during parsing.

To our knowledge these are the first experiments which objectively
demonstrate the utility of punctuation for resolving syntactic
ambiguity and improving parser coverage. They extend work by
Jones (1994) and Briscoe and Carroll (1994) by applying a
wide-coverage text grammar to substantial quantities of
naturally-punctuated text and by quantifying the contribution of
punctuation to ambiguity resolution in a well-defined probabilistic
parse selection model.

Accurate enough parse selection for practical applications will
require a more lexicalised system.  Magerman's (1995) parser is an
extension of the history-based parsing approach developed at IBM
(e.g.\ Black, 1993) in which rules are conditioned on lexical and
other (essentially arbitrary) information available in the parse
history. In future work, we intend to explore a more restricted and
semantically-driven version of this approach in which, firstly,
probabilities are associated with different subcategorisation
possibilities, and secondly, alternative predicate-argument structures
derived from the grammar are ranked probabilistically. However, the
massively increased coverage obtained here by relaxing subcategorisation
constraints underlines the need to acquire accurate and complete
subcategorisation frames in a corpus-driven fashion, before such
constraints can be exploited robustly and effectively with free text.

\section*{References}
\newcommand{\book}[4]{\item #1 #4. {\it #2}. #3.}
\newcommand{\barticle}[7]{\item #1 #7. #2. In #5 eds. {\it #4}. #6:~#3.}
\newcommand{\bparticle}[6]{\item #1 #6. #2. In #4 eds. {\it #3}. #5.}
\newcommand{\boarticle}[5]{\item #1 #5. #2. In {\it #3}. #4.}
\newcommand{\farticle}[6]{\item #1 #6. #2. In #4 eds. {\it #3}:~#5.
Forthcoming.}
\newcommand{\uarticle}[5]{\item #1 #5. #2. In #4 eds. {\it #3}. Forthcoming.}
\newcommand{\jarticle}[6]{\item #1 #6. #2. {\it #3} #4:~#5.}
\newcommand{\particle}[6]{\item #1 #6. #2. In {\it Proceedings of the
#3},~#5. #4.}
\newcommand{\lazyparticle}[5]{\item #1 #5. #2. In {\it Proceedings of the
#3}. #4.}
\newcommand{\lazyjarticle}[4]{\item #1 #4. #2. {\it #3}.}
\newcommand{\lazyfjarticle}[4]{\item #1 #4. #2. {\it #3}. Forthcoming.}

\begin{list}{}
   {\leftmargin 0pt
    \itemindent 0pt
    \itemsep 2pt plus 1pt
    \parsep 2pt plus 1pt}

\particle{Billot, S. and Lang, B.}
         {The structure of shared forests in ambiguous parsing}
         {27th Meeting of Association for Computational Linguistics}
         {Vancouver, Canada}
         {143--151}
         {1989}

\book{Black, E., Garside, R. and Leech, G. (eds.)}
     {Statistically-Driven Computer Grammars of English: The
IBM/ Lancaster Approach}
     {Rodopi, Amsterdam}
     {1993}

\barticle{Briscoe, E.}
         {Prospects for practical parsing of unrestricted text: robust
statistical parsing techniques}
         {97--120}
         {Corpus-based Research into Language}
         {Oostdijk, N \& de Haan, P.}
         {Rodopi, Amsterdam}
         {1994}

\jarticle{Briscoe, E. and Carroll, J.}
     {Generalised probabilistic LR parsing for unification-based grammars}
     {Computational Linguistics}
     {19.1}
     {25--60}
     {1993}

\book{Briscoe, E. and Carroll, J.}
      {Parsing (with) Punctuation}
      {Rank Xerox Research Centre, Grenoble, MLTT-TR-007}
      {1994}

\particle{Briscoe, E., Grover, C., Boguraev, B. and Carroll, J.}
         {A formalism and environment for the development of a large
          grammar of English}
         {10th International Joint Conference on Artificial Intelligence}
         {Milan, Italy}
         {703--708}
         {1987}

\book{Carroll, J.}
     {Practical unification-based parsing of natural language}
     {Cambridge University, Computer Laboratory, TR-314}
     {1993}

\particle{Carroll, J.}
         {Relating complexity to practical performance in parsing with
wide-coverage unification grammars}
         {32nd Meeting of Association for Computational Linguistics}
         {Las Cruces, NM}
         {287--294}
         {1994}

\barticle{Carroll, J. and Grover, C.}
         {The derivation of a large computational lexicon for English
          from LDOCE}
         {117--134}
         {Computational Lexicography for Natural Language Processing}
         {Boguraev, B. and Briscoe, E.}
         {Longman, London}
         {1989}

\particle{Church, K.}
         {A stochastic parts program and noun phrase parser for
unrestricted text}
         {2nd Conference on Applied Natural Language Processing}
         {Austin, Texas}
         {136--143}
         {1988}

\book{Elworthy, D.}
     {Part-of-speech tagging and phrasal tagging}
     {Acquilex-II Working Paper 10, Cambridge University Computer
Laboratory (can be obtained from cide@cup.cam.ac.uk)}
     {1993}

\lazyparticle{Elworthy, D.}
         {Does Baum-Welch re-estimation help taggers?}
         {4th Conf. Applied NLP}
         {Stuttgart, Germany}
         {1994}

\book{Garside, R., Leech, G. and Sampson, G.}
     {Computational analysis of English}
     {Longman, London}
     {1987}

\particle{Grishman, R., Macleod, C. and Meyers, A.} {Comlex syntax: building
a computational lexicon} {International Conference on Computational
Linguistics, COLING-94} {Kyoto, Japan} {268--272} {1994}

\book{Grover, C., Carroll, J. and Briscoe, E.}
     {The Alvey Natural Language Tools Grammar (4th Release)}
     {Cambridge University Computer Laboratory, TR-284}
     {1993}

\lazyparticle{Harrison, P., Abney, S., Black, E., Flickenger, D., Gdaniec,
C., Grishman, R., Hindle, D., Ingria, B., Marcus, M., Santorini, B.
and Strzalkowski, T.}
         {Evaluating syntax performance of parser/grammars of English}
         {Workshop on Evaluating Natural Language Processing Systems}
         {ACL}
         {1991}

\particle{Hindle, D.}
         {Acquiring disambiguation rules from text}
         {27th Annual Meeting of the Association for Computational Linguistics}
         {Vancouver, Canada}
         {118--25}
         {1989}

\book{Jackendoff, R}
     {X-bar Syntax}
     {MIT Press; Cambridge, MA.}
     {1977}

\lazyparticle{Jones, B}
             {Can punctuation help parsing?}
             {Coling94}
             {Kyoto, Japan}
             {1994}

\lazyparticle{Magerman, D.}
          {Statistical decision-tree models for parsing}
          {33rd annul Meeting of the Association for Computational Linguistics}
          {Boston, MA}
          {1995}

\particle{de Marcken, C.}
         {Parsing the LOB corpus}
         {28th Annual Meeting of the Association for Computational Linguistics}
         {New York}
         {243--251}
         {1990}

\book{Nunberg, G.}
     {The linguistics of punctuation}
     {CSLI Lecture Notes 18, Stanford, CA}
     {1990}

\jarticle{Pereira, F. and Warren, D.}
         {Definite clause grammars for language analysis -- a survey
          of the formalism and a comparison with augmented transition
          networks}
         {Artificial Intelligence}
         {13.3}
         {231--278}
         {1980}

\barticle{Sampson, G.}
         {Susanne: a Doomsday book of English grammar}
         {169--188}
         {Corpus-based Research into Language}
         {Oostdijk, N \& de Haan, P.}
         {Rodopi, Amsterdam}
         {1994}

\jarticle{Sampson, G., Haigh, R., and Atwell, E.}
         {Natural language analysis by stochastic optimization: a
          progress report on Project APRIL}
         {Journal of Experimental and Theoretical Artificial Intelligence}
         {1}
         {271--287}
         {1989}

\lazyparticle{Schabes, Y., Roth, M. and Osborne, R.}
         {Parsing of the Wall Street Journal with the inside-outside
algorithm}
         {Meeting of European Association for Computational
Linguistics}
         {Utrecht, The Netherlands}
         {1993}

\book{Taylor, L. and Knowles, G.}
     {Manual of information to accompany the SEC corpus:
the machine-readable corpus of spoken English}
     {University of Lancaster, UK, Ms.}
     {1988}

\end{list}
\end{document}